\documentclass[a4paper,twoside]{article}

\makeatletter
\providecommand{\@listctr}{0}
\makeatother

\usepackage{subcaption}
\usepackage{calc}
\usepackage{amssymb}
\usepackage{booktabs}
\usepackage{amsfonts}
\usepackage{times}
\usepackage{amstext}
\usepackage{amsmath}
\usepackage{amsthm}
\usepackage{multicol}
\usepackage{pslatex}
\usepackage{apalike}
\usepackage[bottom]{footmisc}
\usepackage[utf8]{inputenc}
\usepackage{graphicx}
\usepackage{float}
\usepackage{longtable,lscape}
\usepackage{rotating}
\usepackage{lineno,hyperref}
\usepackage{color,soul}
\usepackage{SCITEPRESS}     

\usepackage{color,soul}

\begin{document}

\title{Banana Ripeness Level Classification using a Simple CNN Model Trained with Real and Synthetic Datasets}

\author{\authorname{Luis E. Chuquimarca\sup{1,2}, Boris X. Vintimilla\sup{1} and Sergio A. Velastin\sup{3,4}}
\affiliation{\sup{1}ESPOL Polytechnic University, ESPOL, CIDIS, Guayaquil, Ecuador}
\affiliation{\sup{2}UPSE Santa Elena Peninsula State University, UPSE, FACSISTEL, La Libertad, Ecuador}
\affiliation{\sup{3}Queen Mary University of London, London, UK}
\affiliation{\sup{4}University Carlos III, Madrid, Spain}
\email{\{lchuquim, boris.vintimilla\}@espol.edu.ec, sergio.velastin@ieee.org}
}


\abstract{The level of ripeness is essential in determining the quality of bananas. To correctly estimate banana maturity, the metrics of international marketing standards need to be considered. However, the process of assessing the maturity of bananas at an industrial level is still carried out using manual methods. The use of CNN models is an attractive tool to solve the problem, but there is a limitation regarding the availability of sufficient data to train these models reliably. On the other hand, in the state-of-the-art, existing CNN models and the available data have reported that the accuracy results are acceptable in identifying banana maturity. For this reason, this work presents the generation of a robust dataset that combines real and synthetic data for different levels of banana ripeness. In addition, it proposes a simple CNN architecture, which is trained with synthetic data and using the transfer learning technique, the model is improved to classify real data, managing to determine the level of maturity of the banana. The proposed CNN model is evaluated with several architectures, then hyper-parameter configurations are varied, and optimizers are used. The results show that the proposed CNN model reaches a high accuracy of 0.917 and a fast execution time.}

\onecolumn \maketitle \normalsize \setcounter{footnote}{0} \vfill

\section{\uppercase{Introduction}}
\label{sec:introduction}

Nowadays, most nutritionists agree that consuming fruits is essential to have a daily nutritious diet. Many people consume several fruits weekly. Markets are the main vendors of fruit, which need to offer their customers high-quality fruit. Therefore, international markets demand quality control of fruits by agro-industries based on international standards~\cite{reid1985product,kader2002us}. One of the parameters for fruit quality inspection is the level of ripeness, which is related to the consumer's appreciation for buying the product and the consumption time of some fruits~\cite{wang2018automatic,bhargava2021fruits}.

The determination of the maturity of the fruit is carried out manually in the agro-industries. There are several weaknesses in the manual method. For example, it is time-consuming, labor-intensive, and can cause inconsistencies in determining banana maturity by the personnel in charge. The rise of machine vision technology together with the evolution of deep learning techniques can overcome the problems mentioned above, potentially being relatively fast, consistent, and accurate.

In agriculture, innovative technologies such as artificial vision are used for various tasks, such as fruit detection, fruit classification, and fruit quality determination (apples, bananas, mangoes, strawberries, blueberries, among others)~\cite{naranjo2020review}. For fruit quality inspection, international standards consider three essential aspects: colorimetry (maturity), geometry (shape and size), and defects (texture). This work focuses on colorimetry that is directly proportional to maturity; that is, depending on the fruit's color level, the maturity level can be identified~\cite{tripathi2021optimized,sun2021improved,cao2021automated,naik2019non}. For example, banana ripeness has seven levels, according to the U.S. state Department of Agriculture (USDA). Bananas are one of the most consumed fruits worldwide due to its good taste and its high level of nutrients.

Convolutional Neural Network (CNN) models are deep learning techniques applied to computer vision to identify banana ripeness. In some works, seven banana maturity levels are considered, but in others, there are only four maturity levels due to the low number of images per level in the datasets~\cite{saragih2021banana}.

Below is an overview of state of the art on CNN models applied to banana maturity:

~\cite{zhu2021support} uses a machine learning technique called Support Vector Machine (SVM) to compare the results with a YOLOv3 model, which is trained on a dataset containing few images of bananas, which generates an inaccurate model. On the other hand, the YOLOv3 model considers only two levels of maturity (semi-ripe and well-ripe) and obtains an accuracy of 90.16\%. Furthermore, the level of maturity of the banana depends on the number of small black areas detected in the texture of the banana, the greater the number of black dots found, the greater the maturity of the banana. However, it does not take into account international standards recommendations. 

~\cite{saragih2021banana} evaluates only two CNN models with the same number of epochs but with a different number of initial layers to identify four banana maturity levels (unripe/green, yellowish green, semi-ripe, and overripe). The evaluation results of the MobileNetV2 and NASNetMobile models showed an accuracy of 96.18\% and 90.84\%, respectively. However, it uses a poor dataset for training and validation, so the models are likely to be inaccurate. Also, it only performs an evaluation with existing models. 

~\cite{ramadhanidentification} used a deep CNN to identify four maturity levels of Cavendish bananas. Banana images are segmented using YOLO and then fed to a VGG16 model trained with two different optimizers: Stochastic Gradient Descent (SGD) and Adam. The optimization model with SGD has a better accuracy of 94.12\% compared to Adam, which has an accuracy of 93.25\%. In that study, the number of images in the dataset is sparse, and more CNN models could have been evaluated.

~\cite{zhang2018deep} designed their CNN model for banana classification considering seven maturity levels. For the training and testing of the CNN model, a dataset generated with a total of 17,312 images of bananas is used. CNN performance results show an accuracy of 95.6\%. However, the research focuses on a single CNN model. In addition, it does not present evaluations of the proposed model against existing models.

After reviewing the state of the art, it can be said that one of the main problems in measuring banana maturity levels is that there are no public image datasets robust enough to work with CNN models. The generation of datasets might include not only real images, but can also take advantage of some software development tools available to generate synthetic images, using for example: Unreal Engine, Unity3D, Dall·e mini, among others.

In this article, a CNN model is proposed to measure four levels of banana maturity, considering that the model is not heavy and simple. Furthermore, this work generates a robust dataset for the four banana maturity levels by using real images plus synthetic images. In the end, the proposed model with the generated dataset is evaluated against existing CNN models, setting specific hyper-parameters. The results obtained from this evaluation verify that the proposed CNN model obtains better metrics than existing CNN models.

This paper is organized as follows. Section \ref{sec:proposed methodology} describes the proposed methodology for developing the work. Section \ref{sec:results} presents the results of banana maturity inspection using the proposed model and evaluates it with existing state-of-the-art CNN models. Finally, the conclusions are given in Section \ref{sec:conclusion}.

\section{\uppercase{Proposed methodology}}\label{sec:proposed methodology}

The contributions of this paper are:

\begin{itemize}
    \item Since there are currently no sufficiently large public datasets with different banana maturity levels, it is proposed to generate an extensive dataset of images that combines synthetic and real data.
    \item An own CNN model to measure banana maturity levels is proposed, the model is simple but with good results.
    \item The proposed CNN model and the generated datasets are evaluated against existing CNN models, using various configurations of the hyper-parameters.
\end{itemize}

The generation of the dataset to be used in the CNN models has two parts: the first is the generation of the synthetic dataset, and the second is the generation of the real dataset. It should be noted that the synthetic dataset is much larger than the real dataset.

The proposed CNN model has two components: the first component is the design and implementation of a simple CNN model called CNN1, which is trained with the synthetic dataset, resulting in the generation of weights. The second component is the application of transfer learning to the same proposed CNN model but with the configuration of weights obtained in the CNN1 model, resulting in a CNN2 model to estimate the maturity levels of the banana, which is trained with a dataset od real images, resulting in a more adjusted CNN model for banana maturity measurement.

\begin{figure}[h]
  \centering
   {\includegraphics[width=7.5cm]{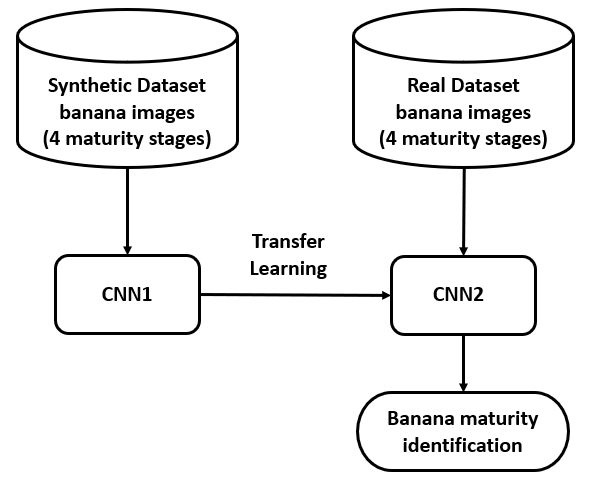}}
  \caption{Banana maturity identification process.}
  \label{fig1}
\end{figure}

Finally, the CNN model is evaluated by comparing it with existing CNN models such as: InceptionV3, ResNet50, Inception-ResNetV2, and VGG19. This evaluation considers the configuration of hyper-parameters and the application of optimizers. For a better understanding of the methodology, see Figure \ref{fig1}.


\subsection{Dataset generation} \label{sec:Dataset generation}

For this work, two types of datasets are generated, one real and the other synthetic, due to the limited number of real images available in public datasets and the time demand in developing real image datasets.

\subsubsection{Real data} \label{sec:Real Data}

\begin{figure}[h]
    \centering
    {\includegraphics[width=7.5cm]{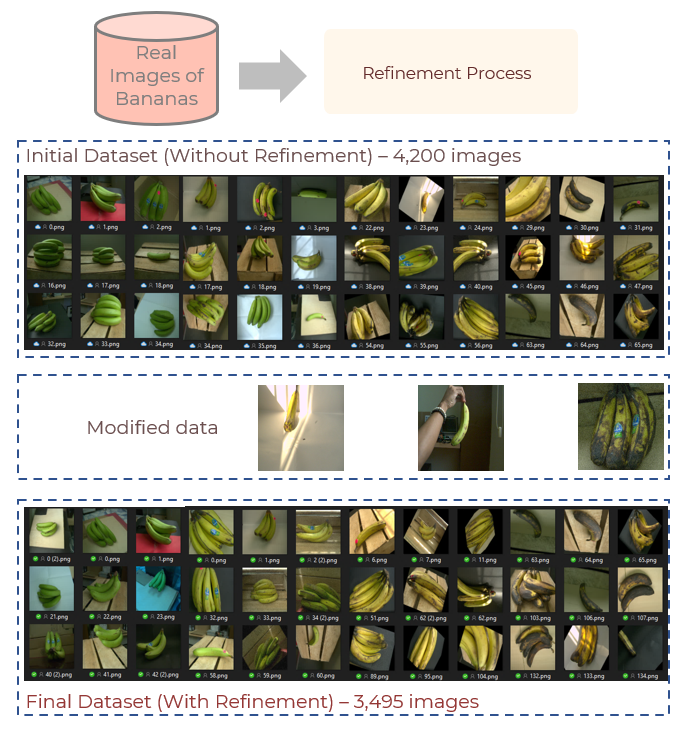}}
    \caption{Real Dataset Refinement.}
    \label{fig2}
\end{figure}

The dataset developed consists of 3,495 real images of Cavendish bananas, which were taken in a laboratory with a climate system between 15°C and 18°C for 28 days (the approximate duration of the ripening period of this type of banana). Also, 4 levels of banana maturity were considered for this work. Therefore, each week the banana passed from one level of maturity to another, as indicated in Table \ref{tab1}. However, bananas have 7 maturity levels, but the number of images that can be acquired per level for the dataset is low. For this reason, it is necessary to group into 4 maturity levels to obtain a greater number of images for the dataset of each banana maturity level. The acquisition of the set of images was carried out daily, considering the maturity cycles of the banana. Therefore, we proceeded to collect 150 images per day. In the end, a total of 4,200 images are generated, of which refinement is performed, removing images with noise, low quality, poor lighting, wrong location of the banana and occlusions (see Figure \ref{fig2}). Therefore, the number of images per level of maturity of the banana will be variable. It should also be mentioned that some of the images in the last days of the last level of maturity are considered rotting and discarded~\cite{ramadhanidentification}.

\begin{table}[h]
    \centering
    \caption{Banana maturity levels per day.}\label{tab1}
    \begin{tabular}{l c c}
        \hline
        \bfseries Duration & \bfseries Level of maturity \\
        \hline
        1 - 6 days              &  A  \\
        7 - 14 days             &  B  \\
        15 - 22 days            &  C  \\
        23 - 28 days            &  D  \\
        \hline
    \end{tabular}
\end{table}

This procedure is costly and tedious because there must be a staff dedicated to the data acquisition process, during the time that the banana ripens. In addition, the conditions in which the bananas are kept must be controlled, such as temperature. Also, it must be taken into consideration that when moving bananas care needs to be taken not to spoil them, because it can cause bruises. So, to obtain a large number of images for the dataset, which CNN models require, the acquisition process must be performed multiple times, leading to high costs and extensive staff time.

The acquisition of the images is carried out in different light conditions and backgrounds. However, a refinement of the images of the real dataset is carried out, considering several aspects, such as whether the bananas are at the corresponding maturity levels, eliminating images where the banana is not clearly defined. Therefore, the final number of images in the real refined dataset is 3,495 images. In the end, data augmentation techniques such as rotation are applied, increasing the amount of data, which is complemented by the generation of synthetic datasets.

There are currently technological tools that allow the generation of synthetic datasets. Therefore, this type of tool is explored, such as the Unreal Engine. It should be noted that there are other types of synthetic image generation engines such as: Unity3D, CARLA or Dall·e mini~\cite{ivanovs2022improving,deiseroth2022logicrank}.

\subsubsection{Synthetic data} \label{sec:Synthetic data}

Synthetic datasets are an important complement to the application in CNN models, due to the low cost and ease of generating large numbers of synthetic images. In addition, there are several CNN models in the literature that make use of domain adaptation and transfer learning techniques by applying synthetic datasets~\cite{charco2021camera}.

\begin{figure}[h]
    \begin{center}
    \includegraphics[scale=0.4, angle = 0]{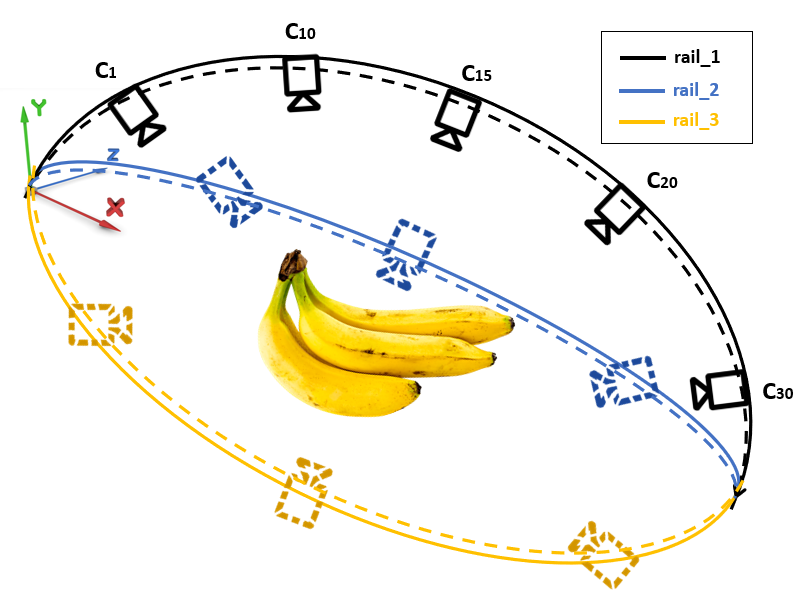}
    \caption{Virtual scenario for the generation of the synthetic images  using Unreal Engine.} \label{fig3}
    \end{center}
\end{figure}

This section describes the process of generating the synthetic banana image dataset using a 3D modeled banana. The Unreal Engine tool is used to create a virtual scenario from which the synthetic data is generated. The virtual environment contains: three rails (rail-1, rail-2 and rail-3), cameras in different positions and angles mounted on each rail (positions C1 - C30), which allow the acquisition of synthetic images of bananas in 3D, as shown in Figure \ref{fig3}. The appearance of the artificial banana ripeness is created using texture from the real images.

Various kinematic components of Unreal Engine are used for the synthetic dataset generation process, such as: Camera Rig Rail (rails), Cine Camera Actor (camera), Level Sequence (sequence), HDRIBackdrop (sky and light), and Material (colors and textures). The camera is fixed to the rail in different positions to capture the synthetic images (see Figure \ref{fig3}). The size of the synthetic images (224x244 pixels) is the same as the real dataset. Therefore, it must be taken into account that the camera's “film back” is configured in millimeters and that it is also proportional to the pixels of the image ($1px = 0.26458333$). The final camera setup is detailed in Table \ref{tab2}. For each camera scan, 30 images are taken on each rail (positions C1 - C30 in Figure \ref{fig3}).

\begin{table}[h]
    \centering
    \caption{Virtual camera parameters for the creation of the synthetic dataset.}\label{tab2}
    \begin{tabular}{l c c}
        \hline
        \bfseries Benefits & \bfseries Value \\
        \hline
        Film back                &  Sensor Width: 59.27 mm  \\
        (Custom)                 &  Sensor Height: 59.27 mm  \\
        \hline
        Lens Settings            &  Min focal length 4.0 mm \\
        (Universal Zoom)         &  Max focal length 1,000 mm   \\
        \hline
        Focus Settings           &  Focus Method: Tracking   \\
                                 &  Tracking focus:\\
                                 &  Actor to track BANANA   \\
        \hline
    \end{tabular}
\end{table}

\begin{figure}[h]
  \centering
   {\includegraphics[width=6.5cm]{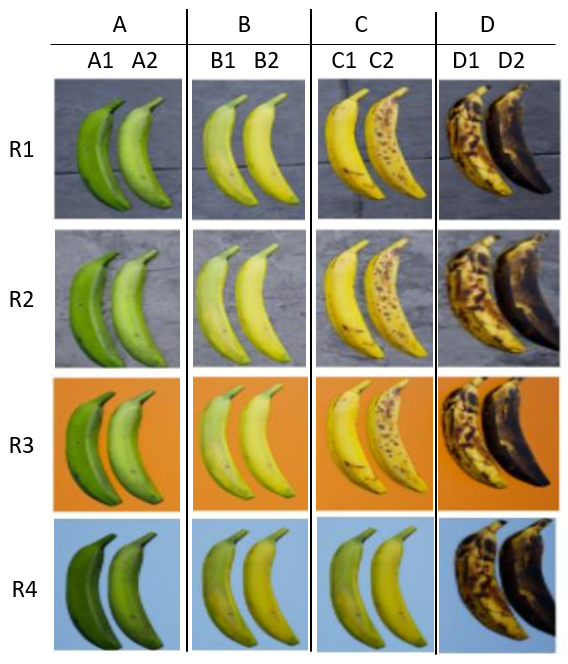}}
  \caption{Synthetic images of banana maturity levels using different backgrounds.}
  \label{fig4}
\end{figure}

Once the virtual scenario is ready, four banana maturity levels are considered for the acquisition of synthetic images as indicated in Figure \ref{fig4} (labels: A, B, C and D). Additionally, two sublevels per maturity level are established and labeled: A1, A2, B1, B2, C1, C2, D1, and D2. In this way, eight colorations are used as shown in Figure \ref{fig4}, modifying the tonality curves and adding spots at maturity levels C and D. For this it was necessary to modify the texture of the banana using Adobe Photoshop CS6 software.

To provide further variability to the virtual scenery, eight different backgrounds were used to capture the synthetic images. The background colors used were: orange, purple, brown and light blue. In addition, materials that come by default in Unreal Engines were used, such as: Asset Platform, Basic Wall, Concrete Tiles (R1) and Rock Marble (R2) (see Figure \ref{fig4}). For the last subclass (level D2), the backgrounds of "Concrete Tiles" were changed to "Ceramic Tile" and "Rock Marble" to "Rock SandStone" to avoid confusion in not distinguish the banana from the background.

\begin{figure}[h]
  \centering
   {\includegraphics[width=6.5cm]{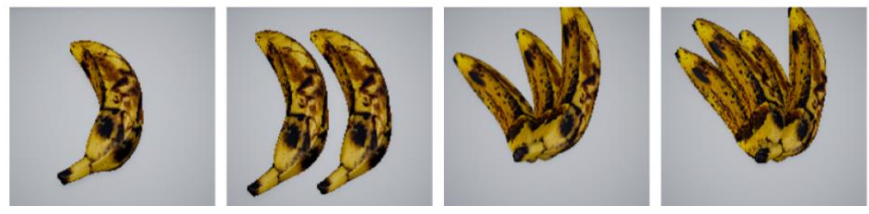}}
  \caption{Variation in the amount of bananas.}
  \label{fig5}
\end{figure}

The number of bananas is alternated between one, two, three, and four to form a bunch, see Figure \ref{fig5}.

In this way, considering the combinations of the proposed scenarios, the total number of synthetic images generated is 161,280, which is approximately 40 times greater than the number of images of the real dataset, which consists of 3,495 images. The number of images per maturity level of both real and synthetic bananas is summarized in Table \ref{tab3}.

\begin{table}[htb]
    \centering
    \caption{Number of images in the banana dataset.}\label{tab3}
    \begin{tabular}{p{1.5cm} p{1.7cm} p{1.5cm}}
        \hline
        \bfseries Banana Maturity Level & \bfseries Number of Synthetic Images & \bfseries Number of Real Images \\
        \hline
        Level A               &  40,320  &  1,429  \\
        Level B               &  40,320  &  815   \\
        Level C               &  40,320  &  559  \\
        Level D               &  40,320  &  692   \\
        \hline
    \end{tabular}
\end{table}

On the other hand, before feeding any CNN with an image dataset, the RGB images must be normalized to obtain good results and to speed up the computational calculations~\cite{sola1997importance}. In addition, the sizes of the images must be standardized, a batch size defined, and the categorical variables encoded in numbers. This last process is applied to both the real and the synthetic image datasets.

\subsection{Description of the proposed model} \label{sec:proposed model}

\begin{figure*}[htb]
    \centering
    \includegraphics[width=1\textwidth]{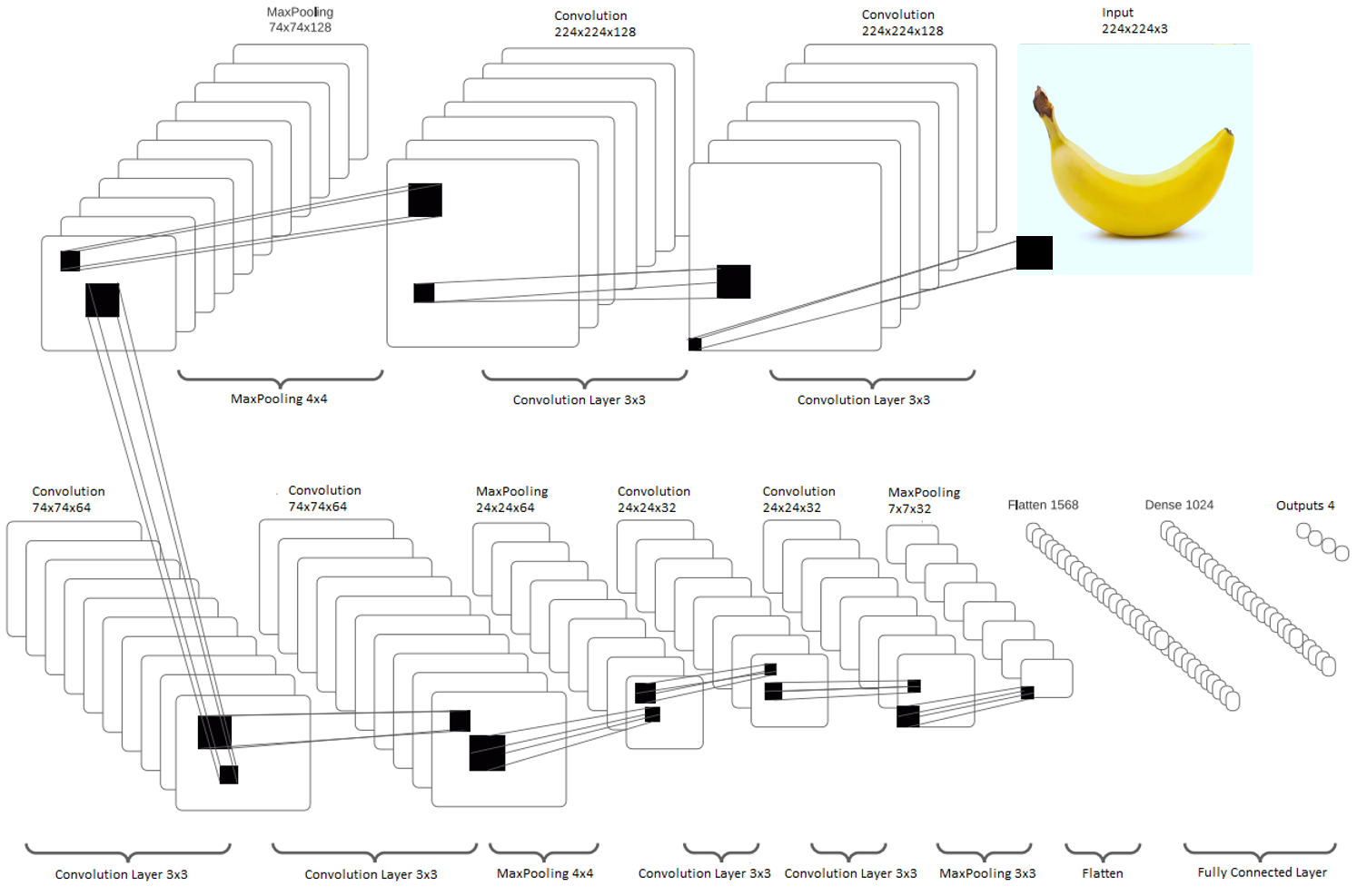}
    \caption{Architecture of the proposed CNN model (CIDIS).}
    \label{fig6}
\end{figure*}

The proposed model (CIDIS) consists of two convolution layers followed by a max pooling layer. This configuration is repeated three times, and fully connected layers follow. The model receives as input images of size 224x224 pixels with a depth of three due to the RGB color channels, and in the end, there are four outputs associated with the four levels of maturity, as shown in Figure \ref{fig6}).

Then, the transfer learning technique is applied using the CIDIS model trained with the synthetic image dataset. This model has stored the scheme and the weights of all its layers. Subsequently, the stored model is loaded into a new instance of the same network, transferring all the weight matrices learned from training with the set of synthetic images. In addition, the last layers of the network (fully connected) had to be removed to apply the optimizers. Finally, the training of the first layers (convolutional and pooling layers) is frozen so that the learned knowledge is not modified, so only the fully connected layers are trained.

When the transfer learning technique is applied to the CIDIS model, this is trained with the real image dataset, which is refined to obtain better results. For this training, the fully connected layers are added, and the optimizer called Adagrad is used because a considerably small dataset is used compared to the synthetic image dataset. This way, the CNN2 model was obtained.

It is important to mention that when the CNN1 model was trained directly with the real image dataset, the accuracy values obtained were lower compared to the CNN2 model, which was fed with the values of CNN1 applying transfer learning, which was trained with the synthetic dataset. 

The CNN2 model is optimized with the following actions such as:

\begin{itemize}
    \item Change the learning rate.
    \item Use dropout layers. 
    \item Change the number of epochs.
    \item Change the batch size value.
    \item Choose between the two proposed optimizers (Nadam and Adagrad).
\end{itemize}

During the training of the CNN2 model, the Nadam and Adagrad optimizers were used. The first is a Nesterov accelerated adaptive moment estimation optimizer that combines ideas from Adam (a stochastic gradient descent method) that uses few computational resources and NAG (Nesterov accelerated gradient), both of which apply to large datasets~\cite{dozat2016incorporating}. On the other hand, the second is an adaptive algorithm that updates the learning rate as the number of learning iterations increases and is more used in small datasets. Both optimizers allowed to converge quickly and efficiently depending on the dataset used.

Adding dropout layers is also considered to reduce overfitting problems. Therefore, by modifying the values of the hyperparameters, it is possible to verify which ones give best results and to build a final robust model, ready to make predictions of banana maturity levels. Ultimately, this optimized model is evaluated with the real image dataset.

\subsection{Evaluation of the proposed model} \label{sec:evaluation model}

For the evaluation of the proposed model, a literature review is carried out from which the best CNN models previusoly reported in the identification of banana maturity are chosen, such as: InceptionV3, ResNet50, Inception-ResNetV2 and VGG19, which are mentioned in ascending order according to the quantity of parameters to train~\cite{faisal2020deep,behera2021maturity,mohapatra2022deep}. 

The VGG19 model within the state-of-the-art review has high performance, high levels of accuracy, and a considerably low training time (less than the InceptionResNetV2 model)~\cite{behera2021maturity}.

ResNet models are designed for double or triple-layer hopping. So skipping layers reduces the disappearing gradient problem. This study uses the 50-layer ResNet-50. Transfer learning and residual learning are applied to optimize network parameters and system development~\cite{helwan2021deep}.

The Inception-ResNetV2 model has 164 layers. It is selected because it obtained a lower percentage of losses compared to other Inception models in the state-of-the-art~\cite{szegedy2017inception}. This model unites two concepts: Inception (reflecting) and Residual Connection (residual connections)~\cite{he2016deep}. In addition, Inception models allow for more efficient computations and increased depth of networks through dimensionality reductions with stacked 1x1 convolutions. Therefore, the model manages to reduce the consumption of computational resources and avoid overfitting~\cite{szegedy2015going}. The InceptionV3 model reduces computational power consumption, being more efficient than the VGGNet and InceptionV1 models~\cite{kurama2020review}. 


The results of the evaluation of the proposed CNN model against the selected CNN models are indicated in the section \ref{sec:results}.

\section{\uppercase{Results}}\label{sec:results}

This section presents and analyzes the results obtained with the generation of the synthetic banana images and with the refinement process using the real image dataset. In addition, the results obtained with the training of the selected CNN models are presented, as well as the application of the transfer learning technique with the final optimizations made to the proposed CIDIS model. For the evaluation of the models in all cases, a dataset distribution of 60\% train, 20\% test, and 20\% validation is used.

Firstly, the selected CNN models are  evaluated without applying the transfer learning methodology, and furthermore, they are trained only with the real dataset. The results obtained with these models are compared with the proposed CIDIS model using the same conditions, Table \ref{tab4} shows these results. The metric used to evaluate the models was accuracy, which calculates the frequency with which the predictions are equal to the proposed labels (0: level A, 1: level B, 2: level C, 3: level D). The time it takes for a CNN model to classify an image was also measured, as well as its total memory weight.

\begin{table*}[htb]
    \centering
    \caption{Comparison of results with CNN models using real data and without transfer learning.}\label{tab4}
    \begin{tabular}{l c c c c}
        \hline
        \bfseries CNN Model & \bfseries Number of Epochs & \bfseries Accuracy & \bfseries Model Weight (Mb) & \bfseries Average time (ms)\\
        \hline
        VGG19              & 100   & 0.562     & 160     & 364  \\
        ResNet-50          & 100   & 0.816     & 200     & 107  \\
        Inception-ResNetV2 & 100   & 0.869     & 1075    & 224  \\
        \textbf{CIDIS (proposed CNN)}             & 100   & \textbf{0.872}     & \textbf{21}      & \textbf{132}  \\
        InceptionV3        & 100   & 0.849     & 187     & \textbf{79}  \\
        \hline
    \end{tabular}
\end{table*}

With the results of Table \ref{tab4} a comparison of the accuracy values and the average classification time is made. Therefore, the CNN model with the best performance is CIDIS, and so this is the CNN model chosen for CNN1. These results were the starting point for this project, establishing a baseline of what can be achieved only with the real data and without carrying out refinement or transfer learning. The results of this test serve to compare the accuracy between CNN1 and CNN2, and to verify if the results obtained by applying the transfer learning technique are viable.

After this, the CIDIS model (as CNN1 model) was trained on the synthetic data and an accuracy of 1.0 was obtained, which is perfect. This ideal result should not be a good reference for the model, because the images of synthetic bananas with different objects, angles and backgrounds are very similar to each other, and therefore the model easily predicts maturity levels. For this reason, the CIDIS model was evaluated with the images of real bananas, obtaining an accuracy of 0.872. This means that although the CIDIS model accurately predicts images of synthetic bananas, it is necessary to train with images of real bananas to better generalize the model.

\begin{figure}[h]
  \centering
   {\includegraphics[width=7.5cm]{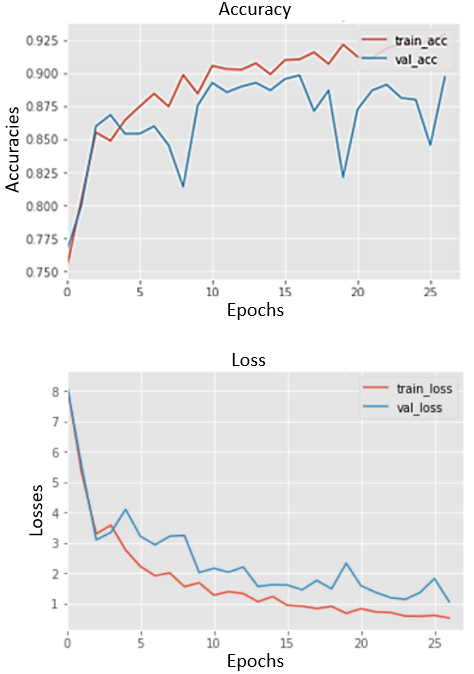}}
  \caption{Results of the accuracy and loss function of the CNN2 model using transfer learning.}
  \label{fig7}
\end{figure}

After the application of the transfer learning technique in the CIDIS model, it is trained with the refined real dataset. For this training, the fully connected layers were added and the Adagrad optimizer was selected, because the real dataset is considerably smaller compared to the synthetic dataset. Figure \ref{fig7} plots the loss and accuracy functions of the CIDIS model as a CNN2 model, with the real dataset.

It can be seen that in the accuracy graph, the model starts learning with an accuracy of 0.74, and continues with an increasing trend until it stabilizes when the validation accuracy does not improve, obtaining an accuracy of 0.9. The graph of the loss has a decreasing trend until it stabilizes, without reaching overfitting. 

With the results obtained from the CNN2 model, optimizations are applied, such as varying the hyperparameters (batch size, learning rate, epochs) of the CNN model, in addition, optimizers are changed and Dropout layers are added. The results can be seen in Table \ref{tab5}.

\begin{table*}[htb]
    \centering
    \caption{Results of the proposed CIDIS model using real/synthetic dataset and transfer learning.}\label{tab5}
    \begin{tabular}{l c c c c c c}
        \hline
        \bfseries CNN Model & \bfseries Optimizer & \bfseries Dropout & \bfseries Learning Rate & \bfseries Batch Size & \bfseries Epochs & \bfseries Accuracy\\
        \hline
              &     Nadam               & 2 (0.2)   & 0.001     & 50     & 45    & 0.881  \\
              &     Nadam               & 1 (0.2)   & 0.001     & 50     & 45    & 0.891  \\
        CIDIS &     Adagrad             & 2 (0.2)   & 0.01      & 50     & 60    & 0.904  \\
              &     \textbf{Adagrad}             & \textbf{1 (0.2)}   & \textbf{0.001}     & \textbf{50}     & \textbf{50}    & \textbf{0.917}  \\
              &     Adam                & 2 (0.2)   & 0.001     & 50     & 80    & 0.916  \\
              &     Adam                & 1 (0.2)   & 0.001     & 50     & 100   & 0.906  \\
        \hline
    \end{tabular}
\end{table*}

\section{\uppercase{Conclusions}}
\label{sec:conclusion}

It was possible to build a dataset of synthetic bananas, which required lower costs and time invested, compared to taking images of real bananas, considering large volumes of data. In this case, generating 4,000 images of real bananas took over 30 days and required multiple people, while generating 161,277 synthetic images took almost as long and was done by a single person using the Unreal Engine software.

A simple own CNN model was implemented to identify banana maturity, it was evaluated with other state-of-the-art CNN models, using a dataset with real images of bananas. Better results were obtained with the new CNN model, which was selected for the development of the proposed work.

In this work, the proposed CNN model (CNN1) was trained with synthetic images, then the transfer learning technique was used to a CNN model called CNN2, which has the same simple architecture as the proposed model. CNN2 was trained and evaluated with a real dataset, obtaining a higher accuracy of 0.917 compared to the proposed CNN model without transfer learning with an accuracy of 0.872. Therefore, it was found that better results are obtained when using the proposed methodology.


\begin{thebibliography}{18}

\bibitem[Behera et~al.(2021)]{behera2021maturity}
Behera, S.~K., Rath, A.~K. and Sethy, P.~K. (2021). Maturity status classification of papaya fruits based on machine learning and transfer learning approach. \emph{Information Processing in Agriculture}, 8(2), 244--250.

\bibitem[Bhargava and Bansal(2021)]{bhargava2021fruits}
Bhargava, A. and Bansal, A. (2021). Fruits and vegetables quality evaluation using computer vision: A review. \emph{Journal of King Saud University-Computer and Information Sciences}, 33(3), 243--257.

\bibitem[Cao et~al.(2021)]{cao2021automated}
Cao, J., Sun, T., Zhang, W., Zhong, M., Huang, B. and Zhou, G. (2021). An automated zizania quality grading method based on deep classification model. \emph{Computers and Electronics in Agriculture}, 183, 106004.

\bibitem[Charco et~al.(2021)]{charco2021camera}
Charco, J.~L., Sappa, A.~D., Vintimilla, B.~X. and Velesaca, H.~O. (2021). Camera pose estimation in multi-view environments: From virtual scenarios to the real world. \emph{Image and Vision Computing}, 110, 104182.

\bibitem[Dozat(2016)]{dozat2016incorporating}
Dozat, T. (2016). Incorporating nesterov momentum into adam.

\bibitem[Faisal et~al.(2020)]{faisal2020deep}
Faisal, M., Albogamy, F., Elgibreen, H., Algabri, M. and Alqershi, F.~A. (2020). Deep learning and computer vision for estimating date fruits type, maturity level, and weight. \emph{IEEE Access}, 8, 206770--206782.

\bibitem[He et al.(2016)]{he2016deep}
He, K., Zhang, X., Ren, S. and Sun, J. (2016). Deep residual learning for image recognition. In \emph{Proceedings of the IEEE Conference on Computer Vision and Pattern Recognition}, 770--778.

\bibitem[Helwan et~al.(2021)]{helwan2021deep}
Helwan, A., Sallam Ma’aitah, M.~K., Abiyev, R.~H., Uzelaltinbulat, S. and Sonyel, B. (2021). Deep learning based on residual networks for automatic sorting of bananas. \emph{Journal of Food Quality}, 2021.

\bibitem[Kader(2002)]{kader2002us}
Kader, A.~A. (2002). US grade standards. In \emph{Postharvest Technology of Horticultural Crops} (pp. 287--300). University of California Agriculture and Natural Resources.

\bibitem[Kurama(2020)]{kurama2020review}
Kurama, V. (2020). A review of popular deep learning architectures: Resnet, inceptionv3, and squeezenet. \emph{Consult. August}, 30.

\bibitem[Mohapatra et al.(2022)]{mohapatra2022deep}
Mohapatra, D., Das, N. and Mohanty, K.~K. (2022). Deep neural network based fruit identification and grading system for precision agriculture. \emph{Proceedings of the Indian National Science Academy}, 1--12.

\bibitem[Naik(2019)]{naik2019non}
Naik, S. (2019). Non-destructive Mango (Mangifera indica L., CV. Kesar) grading using convolutional neural network and support vector machine. In \emph{Proceedings of International Conference on Sustainable Computing in Science, Technology and Management (SUSCOM)}, Amity University Rajasthan, Jaipur-India.

\bibitem[Naranjo-Torres et~al.(2020)]{naranjo2020review}
Naranjo-Torres, J., Mora, M., Hern\'andez-Garc\'ia, R., Barrientos, R.~J., Fredes, C. and Valenzuela, A. (2020). A review of convolutional neural network applied to fruit image processing. \emph{Applied Sciences}, 10(10), 3443.

\bibitem[Tripathi and Maktedar(2021)]{tripathi2021optimized}
Tripathi, M.~K. and Maktedar, D.~D. (2021). Optimized deep learning model for mango grading: Hybridizing lion plus firefly algorithm. \emph{IET Image Processing}.

\bibitem[Ramadhan et~al.()]{ramadhanidentification}
Ramadhan, Y.~A., Djamal, E.~C., Kasyidi, F. and Bon, A.~T. Identification of Cavendish Banana Maturity Using Convolutional Neural Networks.

\bibitem[Reid(1985)]{reid1985product}
Reid, M.~S. (1985). Product maturation and maturity indices. Coop Ext. Univ. of California, Div. of Agric and Natural Resource.

\bibitem[Sola and Sevilla(1997)]{sola1997importance}
Sola, J. and Sevilla, J. (1997). Importance of input data normalization for the application of neural networks to complex industrial problems. \emph{IEEE Transactions on Nuclear Science}, 44(3), 1464--1468.

\bibitem[Saragih and Emanuel(2021)]{saragih2021banana}
Saragih, R.~E. and Emanuel, A.~W.~R. (2021). Banana Ripeness Classification Based on Deep Learning using Convolutional Neural Network. In \emph{2021 3rd East Indonesia Conference on Computer and Information Technology (EIConCIT)} (pp. 85--89). IEEE.

\bibitem[Sun et al.(2021)]{sun2021improved}
Sun, L., Liang, K., Song, Y. and Wang, Y. (2021). An Improved CNN-Based Apple Appearance Quality Classification Method With Small Samples. \emph{IEEE Access}, 9, 68054--68065.

\bibitem[Szegedy et al.(2015)]{szegedy2015going}
Szegedy, C., Liu, W., Jia, Y., Sermanet, P., Reed, S., Anguelov, D., Erhan, D., Vanhoucke, V. and Rabinovich, A. (2015). Going deeper with convolutions. In \emph{Proceedings of the IEEE Conference on Computer Vision and Pattern Recognition}, 1--9.

\bibitem[Szegedy et~al.(2017)]{szegedy2017inception}
Szegedy, C., Ioffe, S., Vanhoucke, V. and Alemi, A.~A. (2017). Inception-v4, inception-resnet and the impact of residual connections on learning. In \emph{Thirty-first AAAI Conference on Artificial Intelligence}.

\bibitem[Wang et~al.(2018)]{wang2018automatic}
Wang, F., Zheng, J., Tian, X., Wang, J., Niu, L. and Feng, W. (2018). An automatic sorting system for fresh white button mushrooms based on image processing. \emph{Computers and Electronics in Agriculture}, 151, 416--425.

\bibitem[Zhang et~al.(2018)]{zhang2018deep}
Zhang, Y., Lian, J., Fan, M. and Zheng, Y. (2018). Deep indicator for fine-grained classification of banana’s ripening stages. \emph{EURASIP Journal on Image and Video Processing}, 2018(1), 1--10.

\bibitem[Zhu and Spachos(2021)]{zhu2021support}
Zhu, L. and Spachos, P. (2021). Support vector machine and YOLO for a mobile food grading system. \emph{Internet of Things}, 13, 100359.

\end{thebibliography}

\bibliographystyle{apalike}

\end{document}